\documentclass[sigconf]{acmart}
\AtBeginDocument{%
}
\copyrightyear{2026}
\acmYear{2026}
\setcopyright{cc}
\setcctype{by-nc-nd}
\acmConference[ICCAD '26]{IEEE/ACM International Conference on Computer-Aided Design}{November 08--12, 2026}{San Jose, CA, USA}
\acmBooktitle{IEEE/ACM International Conference on Computer-Aided Design (ICCAD '26), November 08--12, 2026, San Jose, CA, USA}
\acmDOI{10.1145/3831252.3834165}
\acmISBN{979-8-4007-2873-0/2026/11}

\usepackage{booktabs}
\usepackage{multirow}
\usepackage{colortbl}

\begin{document}
\title{MeltKV: Memory-Aligned KV Cache Compression via Variable-Bit Packed Vector Quantization for Long-Context LLMs}

\author{Jiyun Han}
\affiliation{%
  \institution{Kyung Hee University}
  \city{Yongin}
  \country{Republic of Korea}
}
\email{hjy0013@khu.ac.kr}

\author{Seongjoon Cho}
\affiliation{%
  \institution{Kyung Hee University}
  \city{Yongin}
  \country{Republic of Korea}
}
\email{csj0725@khu.ac.kr}

\author{Seungkyu Choi}
\authornote{Corresponding author.}
\affiliation{%
  \institution{Yonsei University}
  \city{Seoul}
  \country{Republic of Korea}
}
\email{seungkc@yonsei.ac.kr}

\begin{abstract}
Large Language Models (LLMs) increasingly rely on long contexts, making Key-Value (KV) caches dominate memory usage and inference latency. Existing vector quantization (VQ) methods effectively compress KV caches to sub-4-bit levels but often overlook inefficiencies in memory alignment and decoding speed. This work introduces MeltKV, a memory–latency co-optimized KV cache quantization framework for long-context LLM serving. MeltKV employs a memory-aligned variable-bit index packing scheme that adaptively assigns bit-widths to grouped key channels and compactly stores indices within standard data-types, eliminating redundant bits. Additionally, head-wise codebook sharing reduces the total codebook size while maintaining accuracy. Experiments demonstrate that MeltKV effectively compresses KV representations to sub-4-bit levels down to 1 bit and, at comparable perplexity, achieves 48.1\% lower memory usage and 52.9\% lower latency than state-of-the-art VQ methods. 
\end{abstract}




\maketitle

\section{Introduction}
Large Language Models (LLMs) have demonstrated remarkable generalization capabilities across a wide range of tasks, extending their applicability to text generation \cite{generate}, language translation \cite{translate}, and reasoning \cite{reasoning}. As LLMs continue to advance, longer input contexts are increasingly utilized across various domains, enabling applications such as long-document summarization \cite{summary1, summary2} and retrieval-augmented generation \cite{RAG1, RAG2}. To accelerate inference in decoder-only LLMs with long sequence lengths, Key-Value (KV) caching \cite{KVcache} is widely adopted, wherein the key and value activations of all previously processed tokens are stored in memory to avoid redundant recomputation during autoregressive generation.

However, as context lengths grow, the KV cache has emerged as a critical bottleneck for efficient LLM deployment on modern GPUs. The limited memory capacity of current GPUs is easily overwhelmed, as the KV cache can occupy tens of gigabytes under long-context settings. Moreover, frequent access to large KV caches during inference introduces significant memory bandwidth pressure, further limiting throughput. To address these challenges, several quantization-based methods have been proposed to compress the KV cache. For instance, KVQuant \cite{KVQuant} and KIVI \cite{KIVI} employ non-uniform and group-wise quantization schemes, respectively, effectively mitigating outlier effects and achieving 2-bit compression with minimal accuracy degradation. Nevertheless, scalar quantization methods require computationally intensive on-the-fly quantization and dequantization, such as dynamic scaling and grouping, that cannot be preprocessed. Furthermore, scalar quantization inherently struggles to preserve accuracy under aggressive compression due to its limited representational capacity.

Recently, vector quantization (VQ) approaches for KV cache compression \cite{turboquant, CQ, AnTKV} have demonstrated promising results, achieving effective sub-4-bit compression ratios. In VQ, each vector is replaced with its nearest codeword from a predefined codebook, typically constructed via k-means clustering \cite{KmeansPP}. Rather than storing raw values, only the indices corresponding to the codewords are stored. The compression ratio is then determined by the index bit-width and the group size of channels sharing the same codebook. Moreover, the codebook-based structure is well-suited for GPU implementation, making VQ a compelling alternative for KV cache compression.

Despite these advantages, prior works often fail to optimize actual inference performance in terms of both memory efficiency and latency when deploying LLMs with compressed KV caches. Many software-oriented methods prioritize compression ratio alone, assigning distinct codebooks to every attention head and layer without accounting for memory or decoding overhead. Increasing the total number of codebooks introduces additional latency during decompression; for instance, CQ \cite{CQ} reports decoding latency even higher than the FP16 baseline. To mitigate this, MILLION \cite{MILLION} introduces globally shared codebooks to improve decoding speed, but suffers from noticeable accuracy degradation under sub-4-bit compression. Furthermore, aggressive compression schemes often employ non-8-bit indices (e.g., 6–12 bits) to improve accuracy; however, such indices cannot be efficiently aligned with standard data types (e.g., \texttt{uint8}, \texttt{uint16}), resulting in redundant memory consumption. Without specialized data-packing mechanisms, which in turn increase latency, these methods fail to achieve the memory efficiency implied by their nominal bit-widths.

In this work, we propose MeltKV, a KV cache quantization framework that jointly optimizes memory footprint and inference latency by co-designing index bit-width assignment and codebook allocation based on key-cache outlier characteristics.
Unlike prior methods, MeltKV introduces an outlier-aware codebook assignment that groups key channels into codebooks of varying granularities, paired with a memory-aligned index packing scheme.
The resulting indices are compactly stored in standard data types (\texttt{uint8} or \texttt{uint16}), eliminating redundant bits and reducing memory consumption. Additionally, we employ head-wise codebook sharing to reduce total codebook size without compromising accuracy, thereby further enhancing decoding efficiency. As a result, MeltKV achieves comparable or superior perplexity to state-of-the-art (SOTA) VQ-based methods at equivalent effective bit-widths (1-4 bits), while significantly reducing both memory footprint and inference latency.

Our main contributions are summarized as follows:
\begin{itemize}
\item {We propose a joint precision-layout co-design with memory-aligned index packing and head-wise codebook sharing for efficient KV cache compression.}
\item We develop an end-to-end computation pipeline for MeltKV to enable practical and efficient GPU execution.
\item MeltKV achieves superior accuracy across multiple compression levels compared to SOTA KV cache quantization schemes, while also exhibiting the smallest memory footprint and significantly reduced decoding latency.
\end{itemize}

\section{Background}\label{sec:background}
\subsection{LLM Inference and KV Cache}
\label{sec21}
Transformer-based Large Language Models (LLMs) are built upon the self-attention mechanism \cite{Attention}, which enables modeling of contextual dependencies. During inference, the prefill stage computes Key–Value (KV) representations for the entire input sequence, and the decoding stage generates tokens autoregressively. The resulting KV tensors are cached to avoid redundant attention computations, but the cache size grows linearly with sequence length, number of layers, attention heads, and embedding dimensionality.

As shown in Figure \ref{fig1}, decoding must repeatedly load and dequantize the KV cache produced during prefill, introducing substantial memory access and computational overhead. This process becomes the dominant bottleneck in long-context inference and significantly slows generation. While the KV cache is essential for preserving contextual information, it places considerable pressure on GPU memory bandwidth and capacity during large-scale LLM inference. These constraints have driven recent research toward improving KV cache efficiency through quantization and other compression techniques.

\subsection{Vector Quantization for KV Cache}\label{sec22}
To address the memory bottleneck introduced by KV caches during LLM decoding, various quantization techniques have been explored. While scalar quantization is simple and efficient, it fails to capture inter-channel correlations and becomes highly sensitive to KV outliers in ultra-low-bit regimes, leading to severe performance degradation \cite{VQLLM}. This motivates the use of vector quantization (VQ) \cite{VQ_first, PQ_first}, which leverages channel-wise correlations and stores compact indices mapped to pre-trained codebook centroids instead of full-precision KV vectors \cite{PQCache, CQ, Son2025NSNQuant}.

\begin{figure}[t]
\centering
\includegraphics[width=1\linewidth]{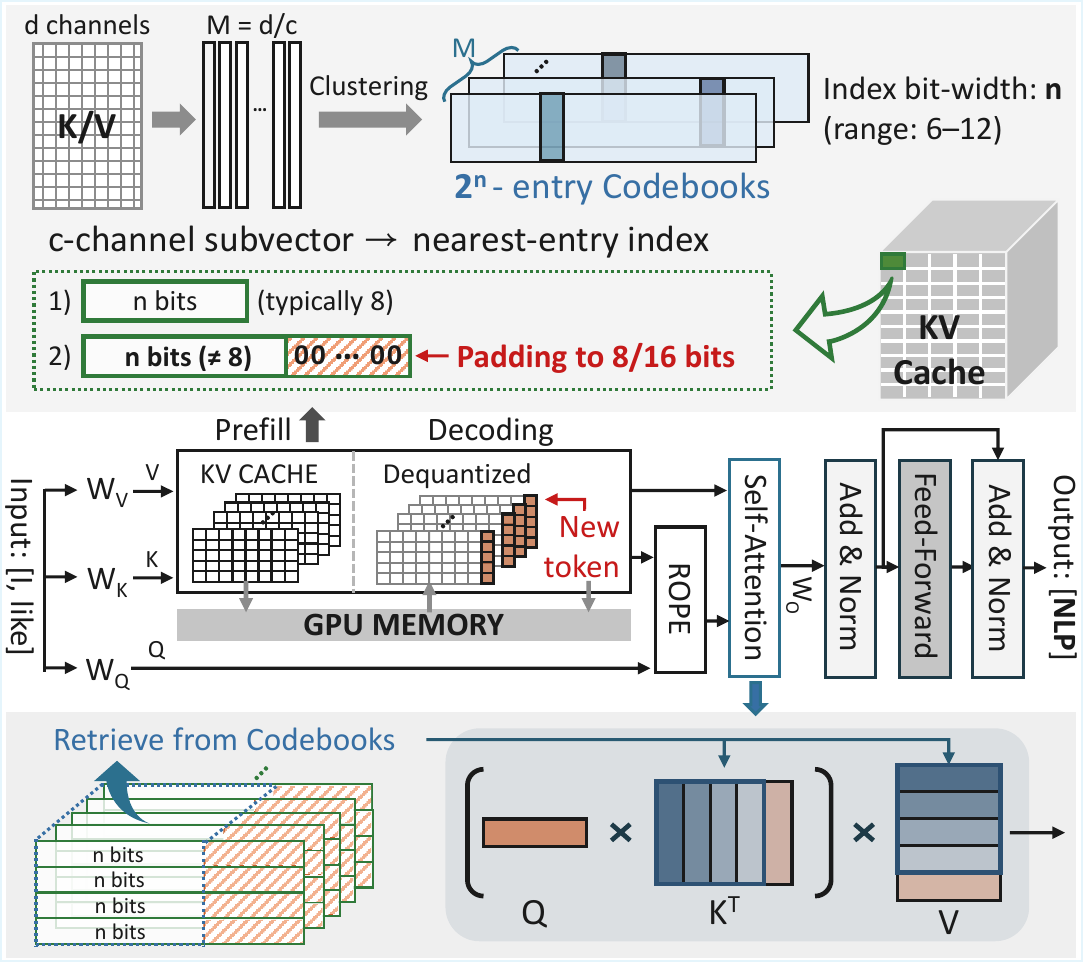}
\vspace{-0.7cm}
\caption{Overview of KV cache vector quantization and its storage trade-offs during prefill and decoding. }
\Description{Overview of vector-quantized KV cache storage and inference.
KV channels are partitioned into subspaces and mapped to codebook indices.
Non-byte-aligned indices require padding to 8- or 16-bit storage.
During decoding, stored indices retrieve codebook entries to reconstruct
keys and values for self-attention.}
\vspace{-0.5cm}
\label{fig1}
\end{figure}

As illustrated in Figure~\ref{fig1}, the channel dimension is partitioned into $M$ subspaces of $c$ consecutive channels, and an independent codebook \(C^{(m)}\) with \(K = 2^n\) centroids is trained for each subspace using KV activations, 
where each subvector is encoded by a codebook index.
Each centroid \(c_k^{(m)}\) is optimized to minimize the within-cluster sum of squared errors via the following k-means objective:
\begin{equation}
    \arg \min_{C^{(m)}} \sum_{k=1}^{K} \sum_{x_j \in C^{(m)}_k} \left\| x_j - c_k^{(m)} \right\|^2
    \label{eq:placeholder_label}
\end{equation} 

Codebooks may be shared globally or instantiated per layer or head, depending on memory capacity and bandwidth constraints. During prefill, each KV vector is divided into $M$ sub-vectors, and every sub-vector \(x^{(m)}\) is mapped to its nearest centroid, and only the resulting $n$-bit index is stored. During decoding, these indices retrieve centroids that are concatenated to reconstruct KV vectors. 

VQ reduces the per-token KV footprint from an FP16 $d$-dimensional vector to roughly 
\(M \times n\) bits of indices, making the choice of $M$ and $n$ central to balancing compression ratio, quantization error, and GPU memory bandwidth. Although index tensors dominate memory usage in long-context settings, codebooks—while relatively small—must be accessed repeatedly from GPU global memory during decoding, introducing additional latency and bandwidth overheads. Consequently, the effectiveness of VQ depends not only on the compression parameters but also on the size and organization of the codebooks, necessitating joint optimization of subspace partitioning, bit-width selection, and codebook configuration.
 
\subparagraph{\textbf{Quantitative Analysis. }}
The memory footprint of an FP16 KV cache can be quantified as:
\begin{equation}
\text{KVSize}_{\text{FP16}}
= B \times T \times l \times h \times d \times 16\,\text{bits} \times 2
\end{equation}
where $B$, $T$, $l$, $h$, and $d$ denote the batch size, context length, number of layers, number of KV attention heads, and per-head KV embedding dimension, respectively. 
The factor of 2 accounts for storing both Key and Value tensors. VQ reduces this footprint by partitioning the embedding dimension into $M=d/c$ subspaces and representing each $c$-dimensional sub-vector with an $n$-bit centroid index drawn from a learned codebook $C^{(m)}$ of size $K=2^n$.

The memory footprint of the VQ-based KV cache is given by
\begin{equation}
\text{KVSize}_{\text{VQ}}^{\text{exact}}
=
\text{KVSize}_{\text{index}}
+
\text{KVSize}_{\text{codebook}},
\end{equation}
where
\begin{equation}
\text{KVSize}_{\text{index}}
=
B \times T \times l \times h \times M \times n \, \text{bits} \times 2,
\end{equation}
\begin{equation}
\text{KVSize}_{\text{codebook}}
=
N_{\text{cb}} \times M \times 2^n \times c \times b_{\text{cb}}.
\end{equation}
Here, $N_{\text{cb}}$ denotes the number of codebooks, and $b_{\text{cb}}$ denotes the bit precision of each codebook entry. Since $\text{KVSize}_{\text{codebook}}$ is a static overhead independent of the sequence length, it becomes negligible in the long-context regime.
Accordingly, for long contexts, the KV-cache compression ratio relative to FP16 is given by
\begin{equation}
\frac{\text{KVSize}_{\text{FP16}}}{\text{KVSize}_{\text{VQ}}}
\approx
\frac{16c}{n}, \;\;\;\;\; effective\; bits=\frac{n}{c}
\end{equation}

Hence, smaller $n$ or larger $c$ yield a higher compression ratio. For instance, a configuration of ($c=4$, $n=8$) corresponds to an effective 2-bit quantization per channel and achieves approximately 87.5\% memory savings 
relative to FP16.

\begin{figure}[t]
\centering
\includegraphics[width=1\linewidth]{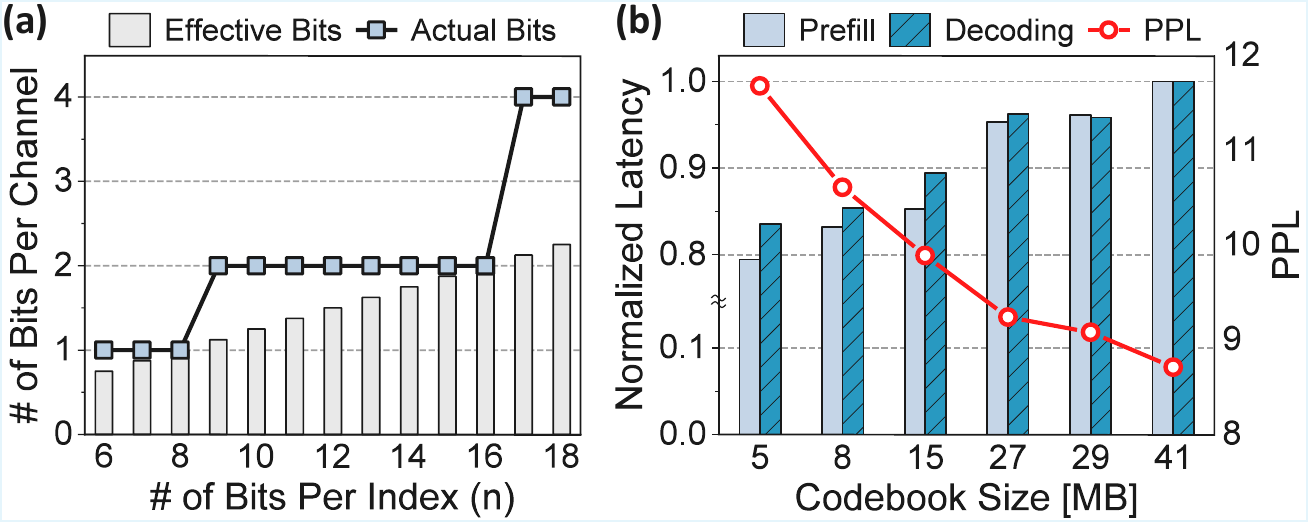}
\vspace{-0.7cm}
\caption{(a) Effective vs. actual VQ bit-widths with padding overhead; (b) Codebook-size trade-off between latency and perplexity (PPL). }
\Description{Two plots illustrating hardware inefficiencies in vector
quantization. The left plot compares effective index bit-width with the
larger actual storage width caused by byte alignment and padding.
The right plot shows that increasing codebook size improves perplexity
but increases both prefill and decoding latency.}
\vspace{-0.1cm}
\label{fig2}
\end{figure}

\subsection{Hardware Inefficiency in VQ}
\label{sec:sec23}
As discussed in Section \ref{sec22}, VQ stores codebook indices rather than raw values, where $n$ denotes the bit-width per subspace index. 
However, common CPU/GPU architectures only support standard data types such as \texttt{uint8}, \texttt{uint16}, and \texttt{uint32} for memory storage. When $n$ does not align with these standard bit-widths, unused bits must be padded, resulting in memory inefficiency. Consequently, traditional VQ methods such as product quantization—widely used in similarity search—typically fix $n$ to 8 so that indices can be stored directly in \texttt{uint8} format. 
In contrast, recent KV quantization approaches \cite{CQ, Son2025NSNQuant, Yao2025VecInfer, VQKV} adopt varying bit-widths to improve the precision–compression trade-off in sub-4-bit regimes, aiming to minimize accuracy degradation. 

As shown in Figure \ref{fig1}, 
when $n$ differs from 8,
non-power-of-two index widths necessitate padding, introducing redundant memory consumption and bandwidth overhead. Figure \ref{fig2}(a) illustrates the discrepancy between the theoretical index bit-width (i.e., the effective bit-width) and the actual bit-width required for storage. For instance, commonly adopted configurations such as $n = 6$ or $n = 12$ in sub-4-bit VQ do not pack efficiently into standard data types, incurring memory overhead beyond what the nominal bit-width implies — for example, storing a 12-bit index in \texttt{uint16} results in a 33\% memory overhead. 
Motivated by this inefficiency, we analyze KV distributions and introduce an index packing scheme designed to align physical memory storage with theoretical compression ratios, while maintaining high accuracy under sub-4-bit quantization.

A second limitation stems from the excessive number of codebooks. Since codebooks occupy only a small fraction of the overall memory footprint relative to indices, prior works often assign distinct codebooks to different attention heads and layers to improve accuracy. However, frequent codebook lookups during decoding incur non-negligible memory overhead, resulting in substantial latency.
Figure \ref{fig2}(b) shows the trade-off between perplexity and prefill/decoding latency under different codebook sizes. Larger codebooks improve fidelity and approach FP16 accuracy, but significantly increase latency, even exceeding the FP16 baseline, as observed in CQ \cite{CQ}. While MILLION \cite{MILLION} shares subspace-specific codebooks globally across layers and heads to reduce latency, this approach fails to maintain accuracy under aggressive sub-4-bit quantization. Therefore, a careful analysis of this trade-off is required to achieve both high accuracy and efficient inference.

\begin{figure}[t]
\centering
\includegraphics[width=1\linewidth]{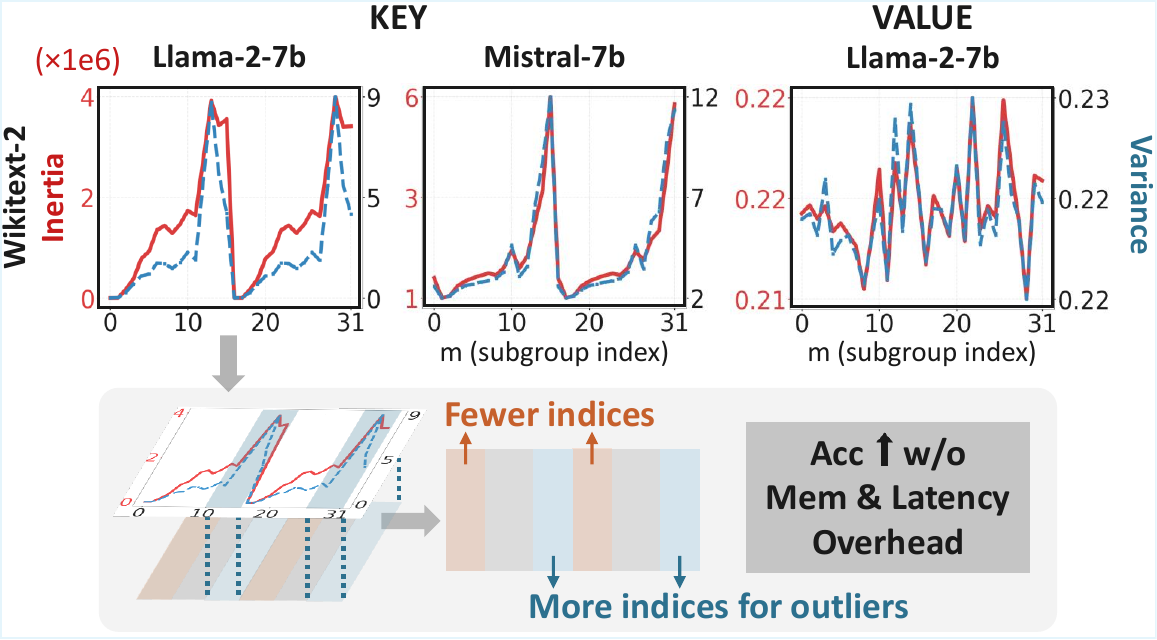}
\vspace{-0.7cm}
\caption{Channel-subspace inertia and variance analysis.} 
\Description{Channel-subspace statistics for key and value tensors.
Key tensors from Llama-2-7B and Mistral-7B show large variation in
clustering inertia and variance across subspaces, whereas value tensors
show relatively small variation. The illustration indicates assigning
more indices to difficult key subspaces and fewer indices to easier ones.}
\vspace{-0.5cm}
\label{fig3}
\end{figure}

\section{MeltKV Methodology}
\label{sec3}
We propose MeltKV, a memory-aligned VQ indexing scheme that assigns different precision levels, corresponding to different codebook sizes across channel groups based on their key outlier characteristics. With a variable-bit index packing mechanism, MeltKV preserves perplexity while reducing memory overhead compared with prior VQ approaches at equivalent compression ratios. Furthermore, MeltKV employs head-wise codebook sharing to reduce overall codebook size and minimize decoding latency. This section provides a detailed description of the proposed method, followed by its hardware-efficient GPU implementation in Section \ref{sec4}.

\begin{figure}[t]
\centering
\includegraphics[width=1\linewidth]{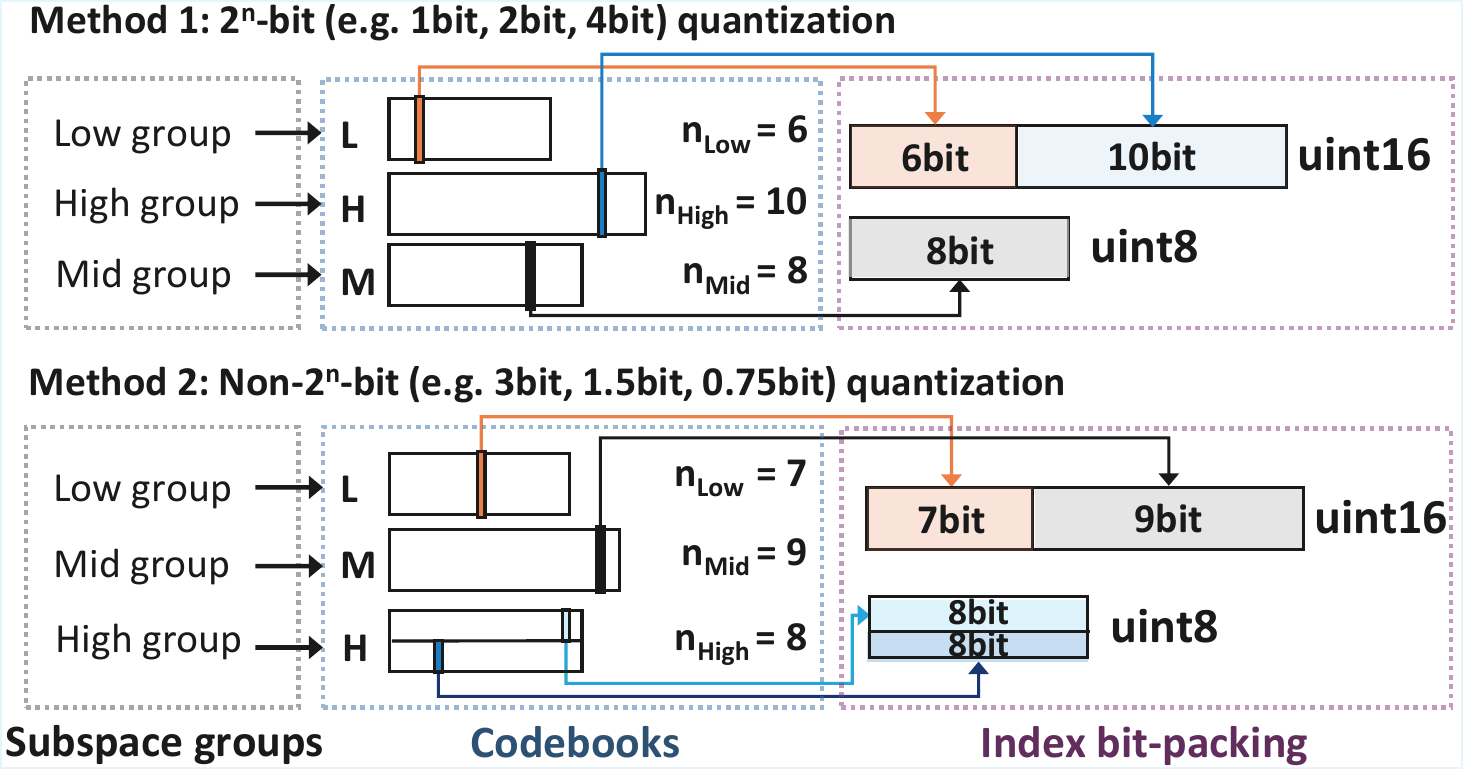}
\vspace{-0.7cm}
\caption{The proposed index bit-packing methods. 
Indices are packed into byte-aligned \texttt{uint8}/\texttt{uint16} formats.}
\Description{Two memory-aligned variable-bit index packing schemes.
For power-of-two effective bit settings, 6-bit and 10-bit key indices
are paired into a 16-bit word while 8-bit indices are stored directly.
For non-power-of-two settings, low- and medium-precision indices are
packed into aligned 16-bit storage, while high-precision groups are
represented using aligned 8-bit indices.}
\vspace{-0.1cm}
\label{fig4}
\end{figure}

\begin{figure}[t]
\centering
\includegraphics[width=1\linewidth]{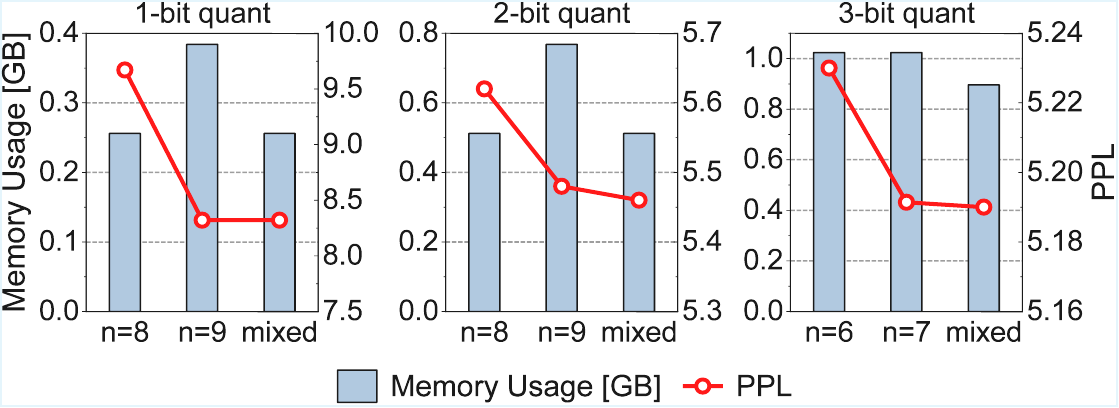}
\vspace{-0.7cm}
\caption{Perplexity vs. index memory usage on Llama-2-7B.}
\Description{Three plots for 1-bit, 2-bit, and 3-bit effective
quantization comparing index memory usage and perplexity across index
configurations. The proposed mixed, memory-aligned configurations
achieve lower perplexity while maintaining or reducing index memory
usage relative to uniform index configurations.}
\vspace{-0.1cm}
\label{fig5}
\end{figure}

\subsection{Precision–Layout Co-Design for Packed KV}
\label{sec31}
The key tensors exhibit substantial channel-wise variation \cite{SKVQ, KIVI} in quantization difficulty, as reflected by the large variance of clustering loss across channel groups in Figure~\ref{fig3}. This heterogeneity is consistently observed across different model families, indicating that a fixed-cardinality codebook \cite{MILLION, CQ} is inherently mismatched to the non-uniform representational demand of key channels. Consequently, easy groups tend to be over-provisioned, while outlier-sensitive groups are under-represented, leading to disproportionate accuracy degradation under aggressive compression.

A straightforward approach is to apply mixed-precision by assigning more bits to difficult groups. However, unlike scalar quantization, heterogeneous precision in VQ introduces a fundamental system-level challenge. Since VQ represents data using centroid indices, varying index bit-widths directly lead to irregular packed layouts. Naively mixing index widths breaks byte/word alignment and introduces padding and storage fragmentation, which increase the effective memory footprint and decoding latency. As a result, the expected accuracy gains from mixed-precision can be offset—or even negated—by degraded memory efficiency and increased decoding latency due to inefficient memory access.

Our key contribution is to make heterogeneous VQ indices compatible with regular packed storage and efficient decoding through joint precision–layout co-design. Channel groups are first categorized into three precision levels — high, medium, and low — based on their k-means clustering loss. The hardware packing unit is then treated as the highest-priority constraint when assigning codebook precision. 8-bit indices are retained as the base decoding primitive, and mixed-precision layouts are constructed to align naturally with \texttt{uint8}/\texttt{uint16} storage, as illustrated in Figure~\ref{fig4}.

Specifically, we restrict the medium-precision index width to 8 bits and pack low- and high-precision indices into a combined 16-bit format, enabling seamless alignment with standard \texttt{uint8} and \texttt{uint16} types. In compression settings with non–power-of-two ratios, the exact compression target can be achieved by extending the low- and medium-level indices to 7 and 9 bits, respectively, while still allowing them to be packed into \texttt{uint16}. For high-variance groups, instead of increasing the index bit-width—which would introduce irregular decoding and higher latency—we partition the subspace into two segments of size $c/2$ with $n=8$ and encode them as two \texttt{uint8} indices to enable fast and regular decoding. In this way, we selectively increase representational capacity only where necessary, while maintaining a regular packed layout and decode-friendly execution. This distinguishes our approach from prior quantization schemes that improve accuracy without considering the resulting padding overhead and system-level inefficiencies.

Unlike key tensors, value distributions show little variance across channels, as shown in Figure \ref{fig3}. Therefore, mixed-precision indexing is applied only to keys, providing better accuracy without additional decoding cost. Figure \ref{fig5} compares perplexity and memory usage under different index configurations for sub-4-bit quantization. The proposed memory-aligned indexing achieves substantially improved perplexity while maintaining the memory footprint of the $n=8$ baseline. In the 3-bit setting where $n=6$ is commonly used, our method yields even lower memory usage with improved perplexity, demonstrating superior indexing efficiency.

\begin{table}[t]
\centering
\captionsetup{skip=4pt}
\caption{Perplexity (↓) sensitivity of KV representations to codebook-sharing granularity under fixed codebook budgets.}
\label{table1}
\scriptsize
\renewcommand{\arraystretch}{1.1} 
\setlength{\tabcolsep}{2.75pt}
\resizebox{1\columnwidth}{!}{  
\begin{tabular}{cccccccccccc}
\toprule
\multicolumn{12}{c}{\cellcolor{gray!10}\textbf{Number of heads per codebook}} \\ 
\multicolumn{6}{c}{\rule{0pt}{2ex}\textbf{Llama-2-7B}} & 
\multicolumn{6}{c}{\rule{0pt}{2ex}\textbf{Llama-3-8B}} \\ 
\multicolumn{3}{c}{\textbf{1-bit quant}} &
\multicolumn{3}{c}{\textbf{2-bit quant}} &
\multicolumn{3}{c}{\textbf{1-bit quant}} &
\multicolumn{3}{c}{\textbf{2-bit quant}} \\ 
\cmidrule(lr){1-6}
\cmidrule(lr){7-12} 
\textbf{Key} & \textbf{Val} & \textbf{PPL} &
\textbf{Key} & \textbf{Val} & \textbf{PPL} &
\textbf{Key} & \textbf{Val} & \textbf{PPL} &
\textbf{Key} & \textbf{Val} & \textbf{PPL} \\ 
1 & 1 & \textbf{{7.18}} & 1 & 1 & \textbf{{5.38}} &
1 & 1 & \textbf{{8.71}} & 1 & 1 & \textbf{{5.99}} \\ 
\hline
1 & 8 & \textbf{{7.11}} & 1 & 8 & \textbf{{5.38}} &
1 & 8 & \textbf{{9.24}} & 1 & 8 & \textbf{{6.01}} \\ 
8 & 1 & \textbf{{8.54}} & 8 & 1 & \textbf{{5.47}} &
8 & 1 & \textbf{{10.74}} & 8 & 1 & \textbf{{6.19}} \\ 
\hline
2 & 4 & \textbf{{7.33}} & 2 & 4 & \textbf{{5.40}} & 2 & 4 & \textbf{{9.70}} & 2 & 4 & \textbf{{6.08}} \\ 
4 & 2 & \textbf{{7.77}} & 4 & 2 & \textbf{{5.42}} & 4 & 2 & \textbf{{10.10}} & 4 & 2 & \textbf{{6.14}} \\ 
\hline 
2 & 8 & \textbf{{7.36}} & 2 & 8 & \textbf{{5.40}} & 2 & 8 & \textbf{{9.89}} & 2 & 8 &  \textbf{{6.09}}\\ 
8 & 2 & \textbf{{8.42}} & 8 & 2 & \textbf{{5.45}} & 8 & 2 & \textbf{{11.04}} & 8 & 2 & \textbf{{6.21}} \\ 
\hline 
8 & 8 & \textbf{{8.32}} & 8 & 8 & \textbf{{5.46}} &
8 & 8 & \textbf{{11.68}} & 8 & 8 & \textbf{{6.22}} \\ 
\bottomrule
\end{tabular}
}
\vspace{-0.5cm}
\end{table}

\begin{figure*}[t]
\centering
\includegraphics[width=1\linewidth]{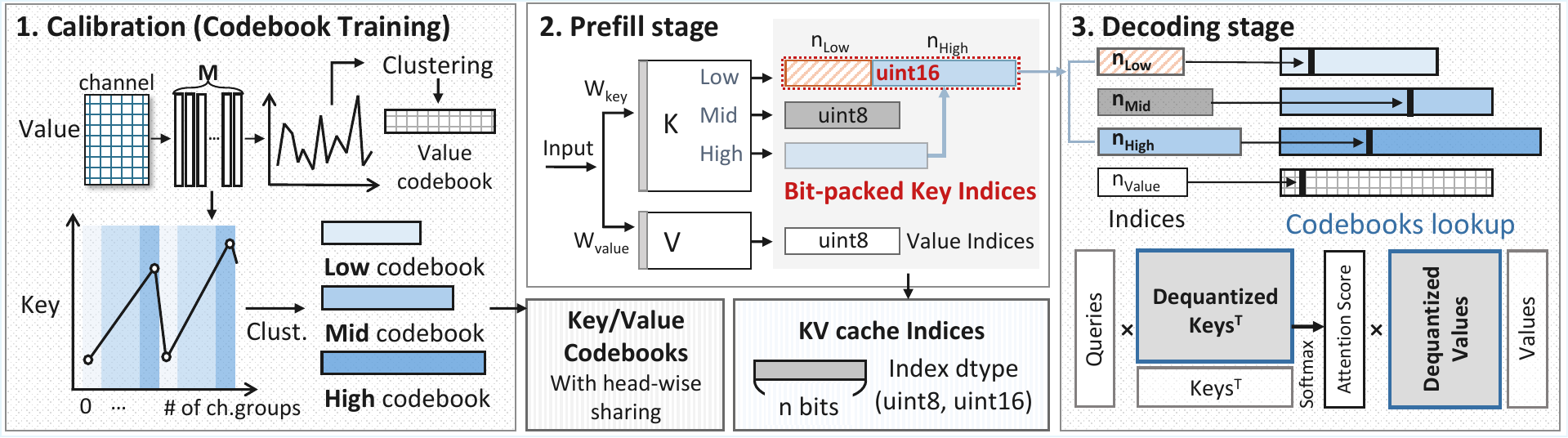}
\vspace{-0.7cm}
\caption{End-to-end computation flow of MeltKV.}
\Description{Three-stage MeltKV processing pipeline. During calibration,
key subspaces are classified into low-, medium-, and high-precision
groups and codebooks are trained with head-wise sharing. During prefill,
key indices are bit-packed into aligned integer formats and value indices
are stored separately. During decoding, indices retrieve codebook entries
to reconstruct keys and values for attention computation.}
\vspace{-0.2cm}
\label{fig6}
\end{figure*}

\subsection{Head-wise Codebook Sharing}\label{sec32}
Extending our statistical observations, we find strong similarity among key and value distributions across heads, while substantial variation exists across layers. Based on this pattern, we adopt head-wise codebook sharing to reduce total codebook size and improve decoding performance without sacrificing accuracy. Table \ref{table1} reports perplexity on Llama models under different sharing granularities.

Key codebooks are more sensitive to sharing in aggressive quantization scenarios (e.g., 1-bit), requiring finer granularity due to their direct influence on $QK^T$, which is the softmax input. In contrast, values exhibit lower variance and do not directly affect attention logits, resulting in negligible degradation under head-level sharing. As a result, value codebooks can be shared more aggressively, reducing memory and accelerating decoding. Notably, for 2-bit quantization, sharing keys and values across up to eight head groups achieves accuracy comparable to independent head- or layer-level codebooks. Even in sub-1-bit compression, stable performance is maintained by adjusting only the number of key-sharing groups.

These findings show that prior VQ methods have overlooked the sensitivity of keys to outliers and the stability of value statistics. Fine-grained codebook construction combined with statistically informed head-wise codebook sharing is essential for maximizing the efficiency of KV cache quantization.    

\begin{figure}[t]
\centering
\includegraphics[width=0.98\linewidth]{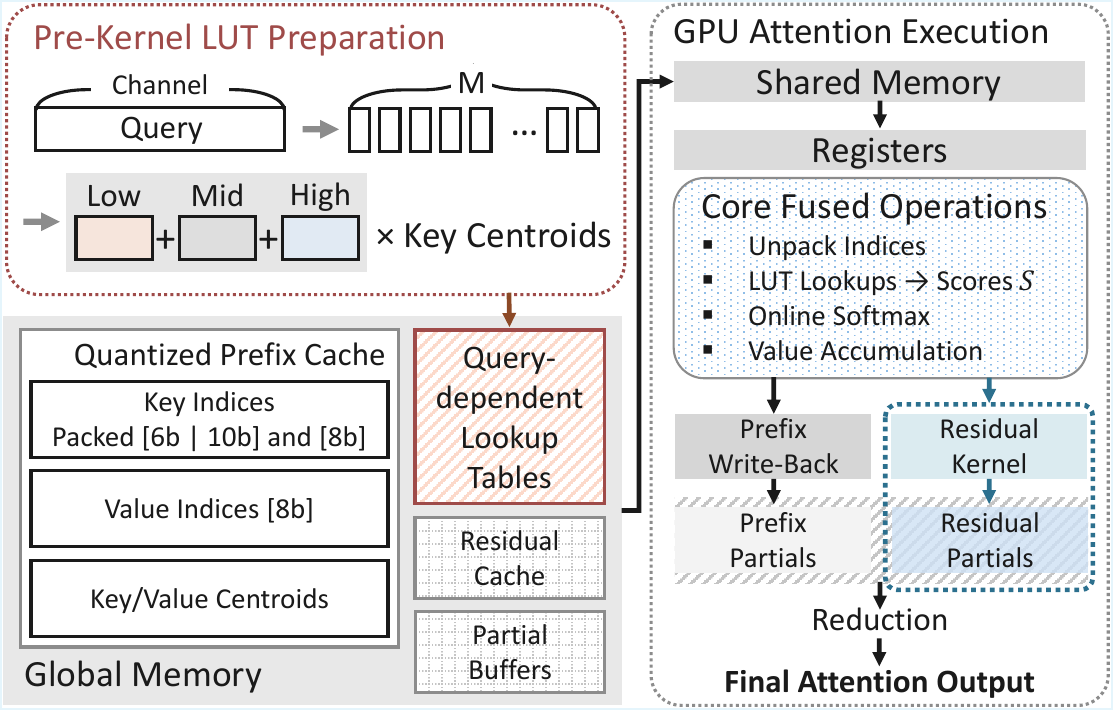}
\vspace{-0.3cm}
\caption{Fused KV Cache Attention on GPU.}
\Description{Dataflow of the fused GPU attention kernel. A query is
partitioned into subspaces and combined with key centroids to prepare
query-dependent lookup tables. Packed key and value indices and centroids
are loaded from global memory, while index unpacking, lookup-table access,
online softmax, and value accumulation are fused on chip. Partial outputs
are reduced with the residual full-precision path to produce the final
attention output.}
\vspace{-0.2cm}
\label{fig7}
\end{figure}

\section{MeltKV Implementation}\label{sec4}
\subsection{Processing Pipeline} 
\label{sec41}
MeltKV is implemented on GPU and comprises three major phases for LLM inference, following the same procedure as general post-training quantization. Figure \ref{fig6} illustrates the overall computation flow of MeltKV. The offline preparation cost is 69\% lower than that of CQ, while MeltKV-specific runtime packing overhead remains below 0.2\% of TTFT. The details of each phase are provided below.

\subparagraph{\textbf{Phase 1: Calibration. }}
Codebooks for keys and values in each subspace are trained using the calibration dataset. A subspace is defined as a group of $c$ adjacent channels, resulting in $M=d/c$ subspaces per head. 
These subspaces are then sorted into three levels from low to high variance and assigned to precision-specific codebooks.
The indexing bit-width and number of groups per level are selected to pack the indices as compactly as possible, as shown in Figure \ref{fig4}. Under this constraint, codebooks are constructed using Fisher-guided centroid learning  \cite{fisher}, where activations with higher Fisher scores are given larger weights during clustering updates.

\subparagraph{\textbf{Phase 2: Prefill. }} 
During the prefill stage, keys and values are generated at each layer, with their size growing linearly with the context length. These generated tensors are encoded into indices and compactly stored in GPU memory until the decoding stage. The proposed variable-bit packing scheme is used to store key indices in \texttt{uint8}/\texttt{uint16} formats while supporting commonly adopted effective bit levels. Value indices are stored using a fixed \texttt{uint8} scale, and $c$ is adjusted to achieve the desired compression ratio. Only for the effective 3-bit configuration of values, $n=6$ is selected with two zero-bit paddings to maintain an equivalent compression ratio.

Existing KV quantization approaches \cite{KVQuant, AnTKV} preserve nearly 1\% of tokens in FP16 to mitigate accuracy degradation in low-precision settings. However, this strategy increases prefill latency by an average of 15.7\%, negatively impacting hardware efficiency. Motivated by the intrinsic attention-sink behavior of LLMs \cite{anchor}, where the first token dominates early attention, we replace percentile-based retention with a single FP16 anchor token, preserving early-stage attention stability with negligible latency ($<$0.1\%).

\subparagraph{\textbf{Phase 3: Decoding. }}
During decoding, VQ indices are mapped to KV tensors via lookup from codebooks stored as lookup tables (LUTs). We employ a customized GPU kernel that fuses dequantization with attention computation to minimize memory traffic and latency. To improve efficiency, indices are pre-reordered to enable more uniform LUT access patterns, and memory accesses are optimized to leverage GPU caching and shared memory. Detailed kernel design is described in Section \ref{sec42}.

\subsection{GPU Kernel Optimization}
\label{sec42}
As shown in Figure~\ref{fig7}, the proposed GPU kernel algorithm consists of two stages: (1) a pre-kernel LUT preparation stage and (2) a GPU attention execution stage for processing the quantized prefix cache. 

First, the input query is decomposed into \(M\) subspaces, and query-dependent LUTs are constructed by multiplying each subspace with the corresponding key centroids. In the current implementation, LUT preparation is performed separately for low-bit, mid-bit, and high-bit centroid groups, and the resulting LUTs are stored in global memory for subsequent access by the split kernel. In addition, the codebooks are shared across heads, which avoids maintaining separate centroid tables for individual heads, thereby reducing codebook storage overhead and amortizing LUT preparation cost across multiple heads.
In the attention execution stage, the split kernel progressively loads compressed key/value indices from global memory into on-chip memory on a tile basis. For key indices, attention scores are computed through LUT-based lookups, whereas for value indices, output vectors are accumulated by referencing the corresponding value centroids. Throughout this process, tile-wise online softmax, as in FlashAttention \cite{FlashAttention2, flash}, is applied to maintain numerical stability over long prefix sequences.

For the quantized prefix, MeltKV avoids materializing dequantized full-precision K/V tensors as intermediate buffers, thereby eliminating redundant data movement and buffer overhead. Instead, compressed indices are decoded on-the-fly within each attention tile, while tile-local indices and reusable intermediate statistics are kept in on-chip memory to reduce global memory traffic and improve data reuse. Partial outputs and log-sum-exp statistics from all splits, together with those from the residual full-precision segment, are then renormalized to produce the final attention output. 

\begin{table}[t]
\centering
\captionsetup{skip=4pt}
\caption{Architectural specifications of the evaluated models. }
\label{tab:model_stats}
\small
\setlength{\tabcolsep}{4.0pt}
\begin{tabular}{lcccc}
\toprule
\textbf{Model} 
& \textbf{\# KV heads} 
& \textbf{\# Layers} 
& \begin{tabular}[c]{@{}c@{}}\textbf{Context}\\\textbf{length}\end{tabular} &
\begin{tabular}[c]{@{}c@{}}\textbf{Hidden}\\\textbf{size}\end{tabular} \\
\midrule
Llama-2-7B              & 32 & 32 & 4k   & 4096 \\
Llama-3-8B              & 8  & 32 & 8k   & 4096 \\
Mistral-7B              & 8  & 32 & 8k   & 4096 \\
Qwen2.5-7B              & 4  & 28 & 131k & 3584 \\
Granite-3.0-8B-Instruct & 8  & 40 & 4k   & 4096 \\
Llama-3-8B-Instruct     & 8  & 32 & 8k   & 4096 \\
\bottomrule
\end{tabular}
\vspace{-0.3cm}
\end{table}

\section{Evaluation}
\subsection{Experimental Setup}\label{sec:settings}

\subparagraph{\textbf{Models.}}~%
We evaluate MeltKV on six representative LLMs from diverse model families: Llama-2-7B-hf \cite{Touvron2023Llama2}, Llama-3-8B \cite{Grattafiori2024Llama3}, Mistral-7B-v0.1 \cite{Jiang2023Mistral7B}, Qwen2.5-7B \cite{qwen2.5}, Granite-3.0-8B-Instruct \cite{granite}, and Llama-3-8B-Instruct. 
As summarized in Table~\ref{tab:model_stats}, these models span both
multi-head attention (MHA) and grouped-query attention (GQA) architectures, with varying numbers of KV heads and layers, hidden sizes, and context lengths.
The evaluated set also includes both base and instruction-tuned checkpoints, as well as short- and long-context models, enabling us to assess whether MeltKV generalizes beyond a single architecture family or deployment regime. 

\subparagraph{\textbf{Tasks.}}~%
Our evaluation covers three settings: perplexity (PPL), zero-shot (ZS) downstream accuracy, and long-context generation \cite{Bai2023LongBench}. We use PPL to measure the impact of KV cache quantization on prefill attention computation. We evaluate zero-shot downstream accuracy on Qwen2.5-7B and report the average over eight reasoning and knowledge benchmarks: ARC-Easy/Challenge \cite{arc}, PIQA \cite{piqa}, SciQ \cite{SciQ}, WinoGrande \cite{winogrande}, HellaSwag \cite{HellaSwag}, OpenBookQA \cite{OpenBookQA}, and MMLU \cite{mmlu}. 
We also evaluate long-context performance on LongBench following the KIVI
\cite{KIVI} protocol and extend the evaluation to six task categories.
Wikitext-2 \cite{wiki} is used for calibration. For fair comparison, we compute attention with full-precision KV during prefill, quantize the resulting KV cache, and use it only during decoding. 

\subparagraph{\textbf{Baselines.}}~%
We compare MeltKV with an uncompressed FP16 KV cache and representative KV cache quantization approaches, including KVQuant \cite{KVQuant} and CQ \cite{CQ}, which represent SOTA scalar and vector quantization methods, respectively. We further compare against MILLION \cite{MILLION}, a hardware-efficient VQ method, to evaluate both accuracy and latency and demonstrate MeltKV’s effectiveness. 
Notably, KVQuant employs non-uniform quantization and retains 1\% of outliers in FP16 to prevent severe accuracy degradation.

\begin{table}[t]
\vspace{0cm}
\centering
\captionsetup{skip=3pt}  
\caption{Perplexity on Wikitext-2 and zero-shot accuracy of various LLMs across different effective bit-level settings. }
\label{tab:perplexity}
\renewcommand{\arraystretch}{1.1}
\setlength{\tabcolsep}{2pt} 
\resizebox{1\columnwidth}{!}{
\begin{tabular}{lccccccc}
\toprule
\multicolumn{2}{c}{\rule{0pt}{3ex}\textbf{}} &
\multicolumn{5}{c}{\rule{0pt}{3ex}\textbf{Perplexity (↓)}} &
\multicolumn{1}{c}{\rule{0pt}{3ex}\textbf{ZS (↑)}} \\ 
\cmidrule(lr){3-7}   
\cmidrule(lr){8-8}   
\multicolumn{1}{c}{\rule{0pt}{3ex}\textbf{Method}} & 
\multicolumn{1}{c}{\rule{0pt}{3ex}\textbf{$\overline{bit}$}} &
\multicolumn{1}{c}{\rule{0pt}{3ex}\textbf{Llama 2}} &
\multicolumn{1}{c}{\rule{0pt}{3ex}\textbf{Llama 3}} &
\multicolumn{1}{c}{\rule{0pt}{3ex}\textbf{Mistral}} &
\multicolumn{1}{c}{\rule{0pt}{3ex}\textbf{Qwen}} &
\multicolumn{1}{c}{\rule{0pt}{3ex}\textbf{Granite}} &
\multicolumn{1}{c}{\rule{0pt}{3ex}\textbf{Avg.}} \\
\cmidrule(lr){1-1}   
\cmidrule(lr){2-2}   
\cmidrule(lr){3-7}
\cmidrule(lr){8-8}
\multirow{1}{*}{Baseline} & 16.0 & 5.12 & 5.54 & 4.76 & 6.13 & 5.17 & 67.7 \\
\hline
KVQuant-1\% & 4.32 & \textbf{5.13} & \textbf{5.57} & 4.78 & \textbf{6.16} & \textbf{5.17} & 67.3 \\
MILLION & 4.00 & 5.33 & 6.56 & 4.89 & 8.33 & 5.32 & 60.2 \\
CQ & 4.00 & 5.14 & \textbf{5.57} & \textbf{4.74} & 6.17 & 5.20 & 67.6 \\
\textbf{MeltKV} & 4.00 & 5.14 & 5.58 & \textbf{4.74} & \textbf{6.16} & 5.18 & \textbf{67.7} \\
\hline
KVQuant-1\% & 3.32 & \textbf{5.19} & 5.68 & 4.84 & 6.27 & \textbf{5.20} & 67.0 \\
MILLION (n=12) & 3.00 & 5.50 & 7.03 & 5.01 & 7.59 & 5.35 & 65.0 \\
MILLION (n=6) & 3.00 & 9.22 & 10.8 & 6.05 & 10.6 & 5.64 & 59.4 \\
CQ & 3.00 & \textbf{5.19} & \textbf{5.67} & 4.80 & 6.26 & 5.32 & \textbf{67.2} \\
\textbf{MeltKV} & 3.00 & \textbf{5.19} & 5.70 & \textbf{4.79} & \textbf{6.25} & 5.24 & 66.9 \\
\hline
KVQuant-1\% & 2.32 & 5.58 & 6.54 & 5.22 & 6.84 & 5.54 & 65.3 \\
MILLION & 2.00 & 15.0 & 42.8 & 20.7 & 26.9 & 6.73 & 55.2 \\
CQ & 2.00 & 5.46 & \textbf{6.08} & 5.02 & 6.68 & 5.58 & 65.6 \\
\textbf{MeltKV} & 2.00 & \textbf{5.40} & 6.09 & \textbf{5.00} & \textbf{6.64} & \textbf{5.39} & \textbf{65.9} \\
\hline
KVQuant-1\% & 1.32 & 93.0 & 45.1 & 14.5 & 40.4 & 12.5 & 50.6 \\
MILLION & 1.00 & 3E+03 & 5E+02 & 1E+03 & 4E+03 & 86.4 & 32.1 \\
CQ & 1.00 & 7.79 & \textbf{9.40} & \textbf{6.67} & \textbf{9.33} & 7.89 & 55.7 \\
\textbf{MeltKV} & 1.00 & \textbf{7.36} & 9.89 & 6.70 & 9.35 & \textbf{7.12} & \textbf{56.9} \\
\bottomrule
\end{tabular}
}
\begin{flushright}
{\footnotesize\textit{
* MeltKV uses 8/8 (K/V) heads sharing a codebook for 3--4-bit and 2/8 for 1--2-bit settings. \\
* Zero-shot (ZS) results for Qwen2.5 are averaged over eight downstream tasks.
}}
\end{flushright}
\vspace{-0.1cm}
\end{table}

\begin{table*}[t]
\centering
\captionsetup{skip=3pt} 
\caption{LongBench evaluation of Llama-3-8B-Instruct with 1--2-bit KV cache quantization under pre-RoPE and post-RoPE \cite{RoPE}.}
\label{tab:longbench}
\renewcommand{\arraystretch}{1}
\setlength{\tabcolsep}{2.9pt}
\resizebox{\textwidth}{!}{
\begin{tabular}{cccccccccccccccc}
\toprule
\multicolumn{2}{c}{\textbf{Method}} & \textbf{Eff. bits} & \textbf{Qasper} & \textbf{QMSum} & \textbf{MNews} & \textbf{TREC} & \textbf{TriviaQA} & \textbf{SAMSum} & \textbf{LCC} & \textbf{RBench} &
\textbf{PsgRetr} & \textbf{NtrvQA} & \textbf{MF-en} & \textbf{Musique} & \textbf{Average(↑)} \\
\cmidrule(lr){1-2}  
\cmidrule(lr){3-3}  
\cmidrule(lr){4-15}  
\cmidrule(lr){16-16} 
\multicolumn{2}{c}{\textbf{Baseline}} & 16.0 & 32.24 & 23.77 & 26.54 & 74.00 & 90.50 & 42.20 & 56.91 & 53.82 & 71.00 & 26.71 & 39.96 & 24.37 & 46.84
\\
\midrule
\multirow{8}{*}{\textbf{pre-RoPE}}
& KVQuant-1\% & 1.32 & 0.912 & 10.50 & 7.558 & 13.50 & 4.471 & 4.890 & \textbf{22.16} & \textbf{24.15} & 4.498 & 0.414 & 3.561 & 0.663 & 8.107 \\
 & MILLION & 1.00 & 5.589 & 10.74 & 7.768 & \textbf{54.00} & 60.60 & 17.17 & 18.25 & 14.28 & 52.12 & 9.264 & 6.772 & 5.857 & 21.87 \\
 & CQ & 1.00 & 6.332 & 19.38 & 17.90 & 35.00 & 82.02 & \textbf{27.48} & 17.39 & 19.82 & 68.53 & 20.16 & \textbf{21.07} & \textbf{15.72} & 29.23 \\
  & \textbf{MeltKV} & 1.00 & \textbf{7.093} & \textbf{19.92} & \textbf{19.51} & 41.50 & \textbf{85.45} & 26.37 & 17.93 & 20.96 & \textbf{69.27} & \textbf{23.03} & 20.50 & 15.21 & \textbf{30.56} \\
\cmidrule(lr){2-16}
 & KVQuant-1\% & 2.32 & 13.24 & 21.11 & \textbf{25.51} & 68.50 & 88.05 & \textbf{40.44} & 40.40 & 32.97 & 13.98 & 18.43 & 32.70 & 12.09 & 33.95 \\
 & MILLION & 2.00 & 26.54 & 22.30 & 24.57 & 70.00 & \textbf{91.44} & 38.58 & 43.54 & 39.43 & \textbf{71.00} & 23.57 & 34.00 & 19.07 & 42.00 \\
 & CQ & 2.00 & 24.65 & 22.27 & 24.65 & 71.00 & 90.52 & 38.84 & 39.37 & 39.90 & 70.00 & \textbf{25.40} & 36.93 & \textbf{23.99} & 42.29 \\
& \textbf{MeltKV} & 2.00 & \textbf{29.44} & \textbf{23.17} & 25.43 & \textbf{73.50} & 90.41 & 40.40 & \textbf{46.54} & \textbf{43.61} & 70.50 & 25.37 & \textbf{40.38} & 22.82 & \textbf{44.30} \\
\midrule
\multirow{3}{*}{\textbf{post-RoPE}}
 & MILLION & 2.00 & 22.43 & 22.56 & 24.23 & 71.00 & 89.73 & 36.80 & 33.03 & 32.20 & 70.09 & 24.27 & 34.03 & 18.42 & 39.90 \\
 & CQ & 2.00 & 23.93 & 21.66 & 24.78 & 68.50 & \textbf{90.97} & 38.43 & 32.73 & \textbf{36.29} & 70.50 & 25.25 & \textbf{37.55} & \textbf{22.22} & 41.07 \\
& \textbf{MeltKV} & 2.00 & \textbf{25.72} & \textbf{22.96} & \textbf{25.25} & \textbf{72.50} & 90.33 & \textbf{39.26} & \textbf{36.64} & 35.57 & \textbf{71.00} & \textbf{26.17} & 36.92 & 21.83 & \textbf{42.01} \\
\bottomrule
\end{tabular}
\vspace{-0.1cm}
}
\end{table*}

\subparagraph{\textbf{{Hardware and Measurement Setup. }}}~%
All experiments are conducted on a single NVIDIA RTX A6000 GPU, and all methods are evaluated under the same hardware and software environment. 
Our implementation is publicly available.\footnote{\url{https://github.com/acalabys/MeltKV}}
Unless otherwise specified, we use batch size 1 and a fixed random seed throughout all experiments. 
We perform warm-up runs before latency measurement.
Time To First Token (TTFT) is defined as the elapsed time from the start of prefill to the generation of the first output token. Decoding latency is reported in ms/token, computed as the average per-token latency over 100 autoregressive decoding steps.

\subsection{Accuracy Analysis}\label{sec:accuracy}
\subparagraph{\textbf{PPL Evaluation.}}~%
Table \ref{tab:perplexity} summarizes the perplexity results under sub-4-bit compression from 4-bit to 1-bit. While prior SOTA methods maintain accuracy at 4-bit, performance degradation becomes evident for KVQuant and MILLION at the 1-bit level. 
In the 3-bit configuration, MILLION requires a large number of centroids ($c=4$, $n=12$) to avoid severe accuracy loss, whereas CQ and MeltKV remain stable with a smaller configuration ($c=2$, $n=6$). 
Although CQ, which provides strong accuracy among prior methods, shows slightly better performance on Llama-3-8B, MeltKV consistently outperforms CQ across most other models and achieves competitive or better-than-SOTA perplexity.  
For 1--2-bit regimes, we improve stability by sharing one codebook per two key heads and allocating more capacity to keys.

\subparagraph{\textbf{Zero-shot Evaluation.}}~%
We evaluate zero-shot downstream accuracy on Qwen2.5-7B, averaged over eight tasks. MeltKV 
achieves the strongest average
zero-shot accuracy across most evaluated settings, with notable gains in the extreme 1-bit regime. This result suggests that MeltKV better preserves task-level generalization under aggressive low-bit compression.

\subparagraph{\textbf{LongBench Evaluation.}}~%
As shown in Table \ref{tab:longbench}, MeltKV 
outperforms existing baselines on average in both 1-bit and 2-bit LongBench evaluations.
Under the 2-bit configuration, it achieves accuracy comparable to FP16 on most tasks, demonstrating strong robustness in long-context settings. For the 2-bit experiments, we assess both pre-RoPE and post-RoPE quantization. Although pre-RoPE generally yields higher accuracy, post-RoPE provides a more decoding-efficient execution by removing the cost of reapplying RoPE during decoding, thereby reducing latency. 
MeltKV supports both approaches and delivers superior average performance in each case, 
enabling practitioners to flexibly balance the accuracy–efficiency trade-off according to deployment requirements. 
We also note that all methods use a residual length of 128 during decoding \cite{zipcache,Yao2025VecInfer,gear}. 

\subsection{System Performance Analysis} \label{sec:decoding}
\begin{figure}[t]
\centering
\includegraphics[width=1\linewidth]{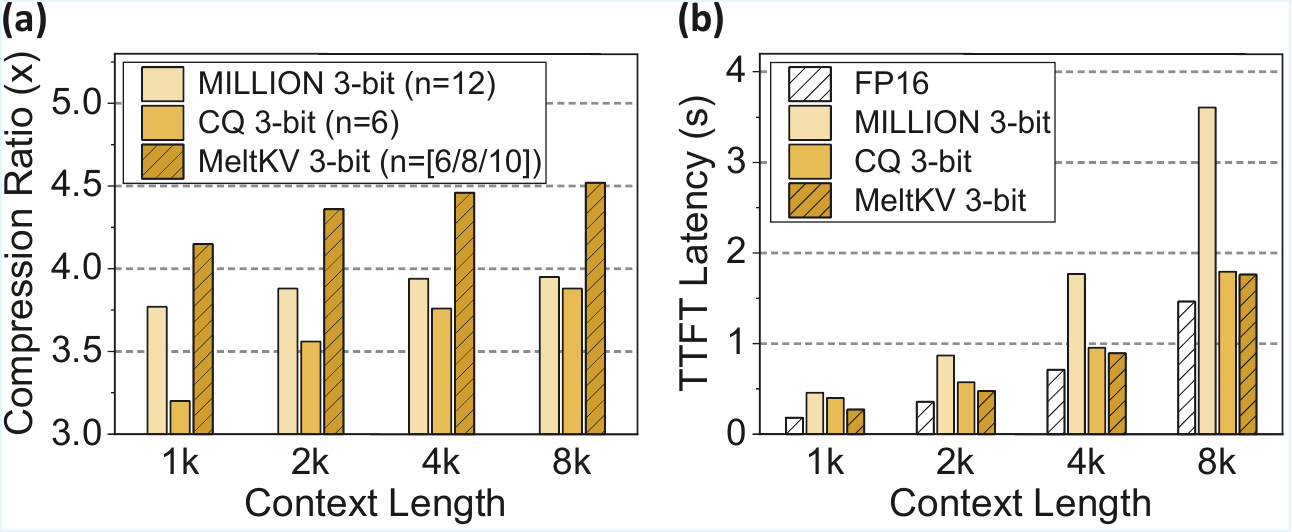}
\vspace{-0.7cm}
\caption{On Llama-3-8B under 3-bit quantization, (a) KV cache compression ratio over FP16, and (b) prefill latency.}
\Description{Two plots comparing 3-bit KV cache implementations on
Llama-3-8B from 1K to 8K context length. MeltKV provides the highest
compression ratio across all context lengths. In the prefill-latency
plot, MeltKV has the lowest TTFT among the evaluated 3-bit quantized
methods, while MILLION has the highest latency.}
\vspace{-0.3cm}
\label{fig8}
\end{figure}

\subparagraph{\textbf{Performance on Different VQ Settings. }}~%
Figure~\ref{fig8} presents the measured memory compression ratio and TTFT latency under the same 3-bit quantization setting with varying VQ configurations. As shown in Figure~\ref{fig8}(a), MeltKV achieves the highest compression ratio — up to 4.5$\times$ across all context lengths — compared to MILLION and CQ, which use $n=12$ and $n=6$, respectively. Values of $n$ not aligned to 8 or 16 bits incur additional storage overhead due to zero-padding. Large $n$ also increases TTFT latency; as shown in Figure~\ref{fig8}(b), MILLION exhibits the highest latency across all context lengths due to slow codebook lookups. Although MILLION employs hardware-efficient global codebook sharing, it requires larger $n$ to match the accuracy of other methods, resulting in both storage and latency inefficiencies. CQ presents a different trade-off: accuracy is improved by employing a large number of codebooks, but at the cost of substantial codebook storage overhead. MeltKV avoids both of these overheads by co-designing quantization with a storage-aligned packed format, thereby achieving the smallest memory footprint and lowest TTFT among 3-bit methods while maintaining SOTA accuracy.

\subparagraph{\textbf{Latency Analysis. }}~%
We evaluate the latency of KV-compressed LLMs under 2-bit quantization, where all VQ baselines use $n=8$, alongside SOTA scalar quantization methods. 
As expected, KV cache quantization introduces additional on-the-fly compression during the prefill phase, leading to higher TTFT compared to the FP16 baseline. Although these approaches significantly reduce the KV cache memory footprint and are therefore essential for long-context LLM serving, excessive latency overhead can limit their hardware efficiency.

\begin{figure}[t]
\centering
\includegraphics[width=1\linewidth]{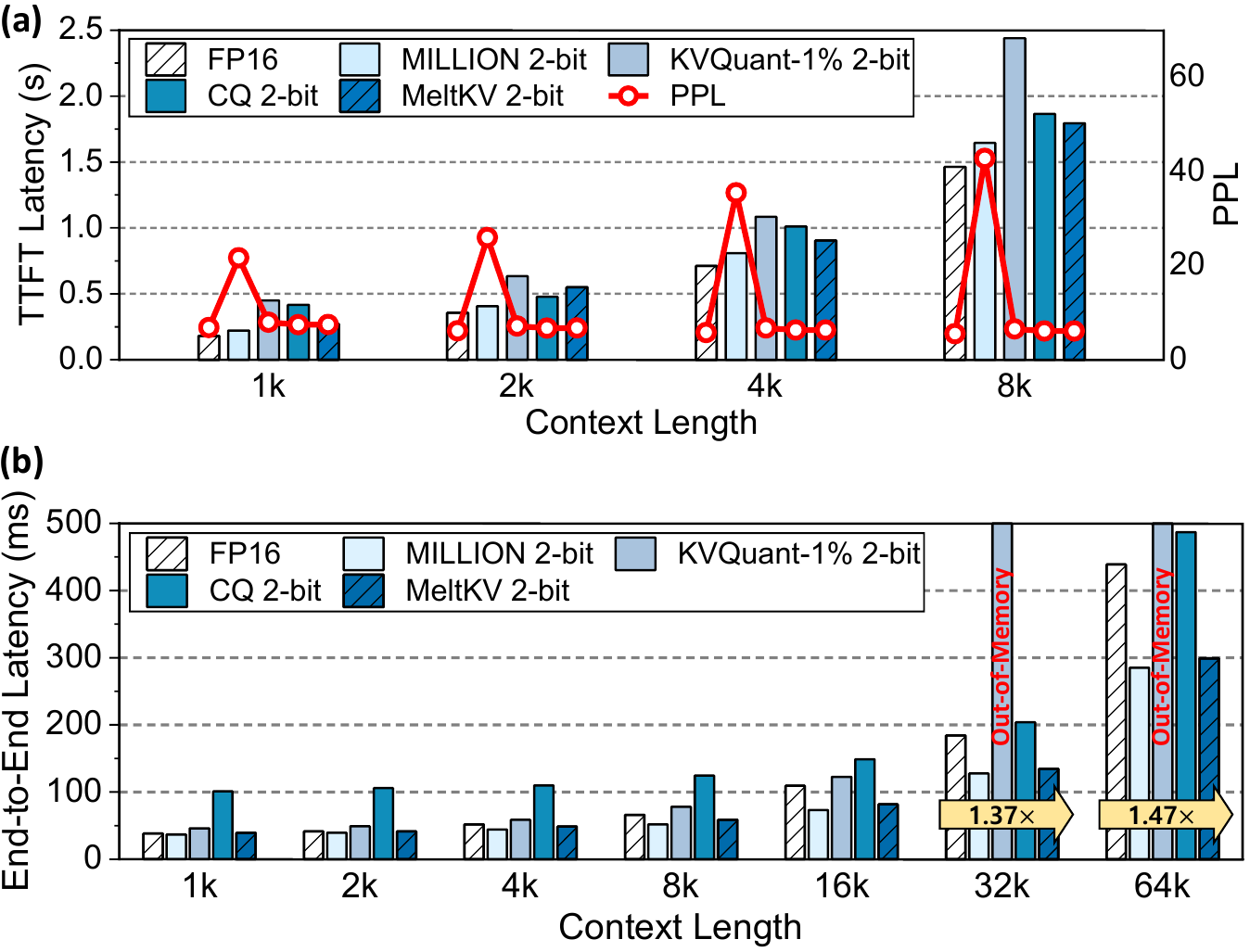}
\vspace{-0.65cm}
\caption{Latency and PPL comparisons on Llama-3-8B-Instruct under 2-bit KV quantization across context lengths: (a) prefill latency and PPL, and (b) end-to-end latency.}
\Description{Two latency comparisons for 2-bit KV cache quantization on Llama-3-8B-Instruct. The first plot compares TTFT and perplexity from 1K to 8K context length and shows severe perplexity degradation for MILLION despite its low latency. The second plot compares end-to-end latency from 1K to 64K context length; MeltKV remains below practical baselines as context length increases, while KVQuant runs out of memory at long context lengths.}
\label{fig9}
\vspace{-0.1cm}
\end{figure}

While MILLION shows poor TTFT latency at 3-bit due to large $n$, its 2-bit setting achieves the lowest latency using $n=8$ and global codebook sharing. However, as shown in Figure~\ref{fig9}(a), this configuration suffers severe PPL degradation. Since MeltKV uses compact codebooks with only 34.4\% of CQ's size, it provides lower TTFT while preserving model accuracy. KVQuant exhibits the highest TTFT due to its additional 1\% outlier-handling path, increasing TTFT by up to 148\% over FP16 and 64.6\% over MeltKV. Overall, MeltKV achieves the lowest TTFT among quantized methods that preserve accuracy within an acceptable range.
Memory-aligned packing reduces KV memory by 20\% without latency overhead, while kernel fusion provides a $2.8\times$ speedup over the same unfused configuration.

\begin{figure}[t]
\centering
\includegraphics[width=1\linewidth]{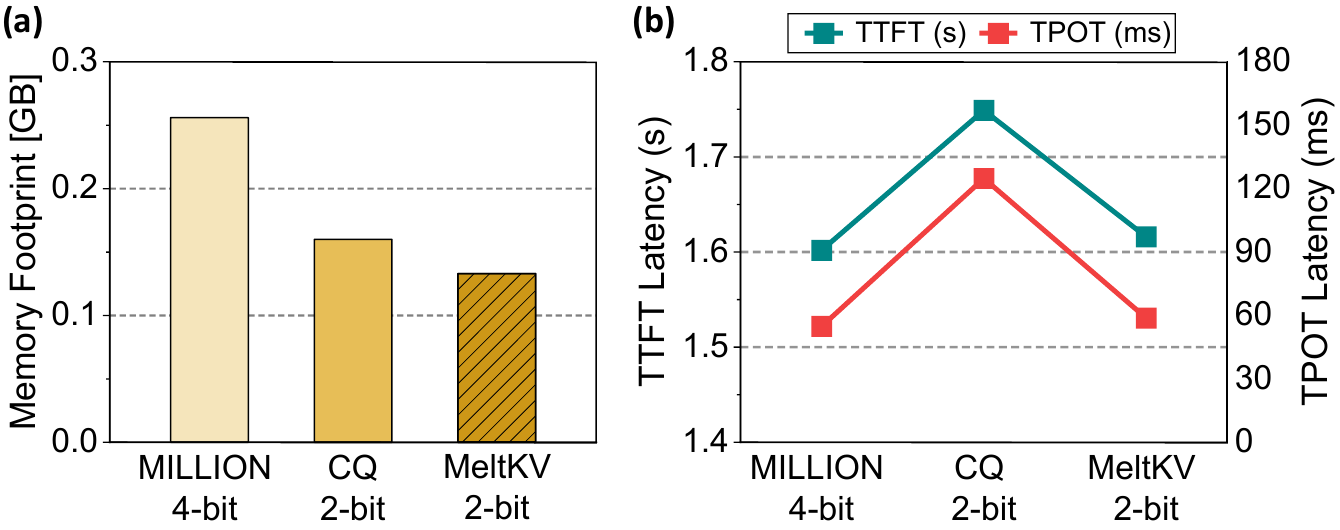}
\vspace{-0.6cm}
\caption{Memory--latency trade-off on Llama-3-8B at 8K under comparable PPL: (a) memory footprint and (b) TTFT and Time Per Output Token (TPOT) latency. }
\Description{Two plots comparing memory and latency at 8K context length under comparable perplexity. The memory plot compares MILLION at 4-bit, CQ at 2-bit, and MeltKV at 2-bit, with MeltKV using substantially less memory than MILLION. The latency plot compares TTFT and time per output token, showing MeltKV with lower latency than CQ while maintaining the smaller 2-bit representation.}
\vspace{-0.1cm}
\label{fig10}
\end{figure}
 
Prefill is performed only once per sequence, whereas decoding dominates serving time, especially for long contexts. Figure~\ref{fig9}(b) shows that MeltKV's decoding speedup outweighs its prefill overhead, resulting in superior end-to-end latency in long-context serving.
This advantage becomes more pronounced as the context length increases: at 64K, MeltKV achieves $1.47\times$ and $1.63\times$ lower latency than FP16 and CQ, respectively; at 16K, it achieves $1.5\times$ lower latency than KVQuant-1\%.

MILLION achieves the fastest decoding latency under 2-bit compression due to global codebook sharing, but this comes at the cost of severe accuracy degradation, limiting its practical applicability. In contrast, CQ exhibits the slowest decoding speed due to its fine-grained codebook design, which requires many separate codebooks across heads and layers. KVQuant-1\% suffers from significantly higher latency and fails to scale to long contexts (OOM at 32K).

\subparagraph{\textbf{SOTA Comparison at Comparable Accuracy. }}~%
\label{sec:comparison}
The core objective of KV cache quantization is to reduce memory while preserving model accuracy. Figure~\ref{fig10} depicts the memory--latency trade-off at 8K with accuracy maintained within approximately +1 PPL of the FP16 baseline. Unlike the fixed-bit comparisons in Figures~\ref{fig8} and~\ref{fig9}, we align configurations (e.g., CQ 2-bit, MILLION 4-bit, and MeltKV 2-bit) that achieve comparable PPL. MILLION uses 4-bit because its 2-bit setting suffers severe PPL degradation. In this accuracy region, CQ prioritizes accuracy, whereas MILLION achieves strong memory and latency efficiency at the cost of reduced accuracy, while MeltKV achieves a more favorable trade-off.

Specifically, MeltKV reduces memory footprint by 48.05\% compared to MILLION, while simultaneously lowering prefill and decoding latency by 2.21\% and 52.91\%, respectively, compared to CQ. 
Moreover, under the same 3-bit compression setting shown in Figure~\ref{fig8}(a), MeltKV reduces memory footprint by up to 23.02\% by eliminating redundant padding bits and aligning the effective bit-width with the actual storage representation. This advantage becomes more pronounced at longer context lengths, yielding savings of up to 2 GB at a context length of 512K tokens. Taken together, these results demonstrate that MeltKV delivers a balanced quantization strategy that achieves a superior memory–latency trade-off at comparable accuracy, offering clear practical advantages over both CQ and MILLION for long-context LLM inference.

\begin{table}[t]
\centering
\caption{Downstream performance by calibration dataset.}
\label{tab:calibration_dataset}
\renewcommand{\arraystretch}{0.95}
\setlength{\tabcolsep}{4pt}
\vspace{-0.2cm}
\resizebox{1\columnwidth}{!}{
\begin{tabular}{lccccccc}
\toprule
\multirow{2}{*}{\textbf{Tasks}} &
\multirow{2}{*}{\textbf{Baseline}} &
\multicolumn{3}{c}{\textbf{MeltKV-3bit}} &
\multicolumn{3}{c}{\textbf{MeltKV-2bit}} \\
\cmidrule(lr){3-5} \cmidrule(lr){6-8}
& & \textbf{Wiki2} & \textbf{C4} & \textbf{PTB} & \textbf{Wiki2} & \textbf{C4} & \textbf{PTB} \\
\midrule
ARC\_c & 43.2 & 42.2 & 42.2 & 43.2 & 41.1 & 40.4 & 40.8 \\
ARC\_e & 75.5 & 75.9 & 75.8 & 75.0 & 73.6 & 74.2 & 73.4 \\
PIQA & 78.1 & 78.0 & 78.2 & 78.0 & 77.7 & 77.6 & 77.4 \\
SciQ & 94.8 & 94.7 & 95.1 & 94.8 & 94.0 & 93.8 & 93.4 \\
WinoG & 69.4 & 69.6 & 69.4 & 69.5 & 67.6 & 66.7 & 67.8 \\
HellaSwag & 57.1 & 57.1 & 57.0 & 56.9 & 55.1 & 55.3 & 54.7 \\
OBQA & 33.2 & 33.0 & 32.8 & 32.4 & 32.0 & 32.8 & 32.4 \\
\midrule
Average & 64.5 & 64.4 & 64.4 & 64.3 & 63.0 & 63.0 & 62.9 \\
\bottomrule
\vspace{-0.5cm}
\end{tabular}
}
\end{table}

\subsection{Calibration Dataset Sensitivity}\label{sec:calibration} 
The impact of different calibration datasets on MeltKV is presented in Table \ref{tab:calibration_dataset}. We construct codebooks using different datasets, including WikiText-2 \cite{wiki}, C4 \cite{C4}, and PTB \cite{ptb}, and evaluate their performance on downstream zero-shot tasks. Across all settings, the average performance difference remains within approximately 0.1 accuracy points, indicating that the choice of calibration dataset has a negligible effect on downstream performance.
In addition, calibration is performed as a lightweight, one-time preprocessing step, resulting in low overhead in practice.

\section{Conclusion}
We proposed MeltKV, a VQ-driven KV cache quantization framework which jointly optimizes memory footprint and latency for long-context LLMs. Through the precision-layout co-design with memory-aligned index packing and head-wise codebook sharing, MeltKV achieves efficient runtime execution while preserving accuracy. 
Experiments show MeltKV outperforms prior approaches in accuracy and hardware efficiency even in ultra-low-bit regimes, offering a practical solution for long-context LLM deployment.

\begin{acks}
This work was partly supported by the National Research Foundation of Korea (NRF) grant funded by the Korea government (MSIT) (RS-2024-00352468) and the Institute of Information \& Communications Technology Planning \& Evaluation (IITP) grant funded by the Korea government (MSIT) (RS-2025-02263669, Development of server-level DRAM-stacked PIM solution for accelerating ultra-large AI models).
\end{acks}

\bibliographystyle{ACM-Reference-Format}
\bibliography{iccad2026_main}
\end{document}